\pdfoutput=1

\documentclass[11pt]{article}

\usepackage{acl}
\usepackage{hyperref}
\usepackage{times}
\usepackage{latexsym}
\usepackage{multirow}
\usepackage{booktabs}
\usepackage{microtype}
\usepackage{multirow}
\usepackage{latexsym}
\usepackage{booktabs}
\usepackage{courier}
\usepackage{amsmath}
\usepackage{booktabs} 
\usepackage{siunitx} 
\usepackage{subfig}
\usepackage{placeins}
\usepackage{tikz}
\usepackage{hyperref}
\usepackage{cleveref}

\usepackage{pdfpages}

\usepackage[T1]{fontenc}

\usepackage[utf8]{inputenc}

\usepackage{microtype}

\usepackage{inconsolata}
\usepackage{}

\title{DFKI-NLP at SemEval-2024 Task 2: Towards Robust LLMs Using Data Perturbations and MinMax Training}

\author{Bhuvanesh Verma \\
  DFKI GmbH \\
  Universität Potsdam \\
  \texttt{bhuvanesh.verma@dfki.de} \\\And
  Lisa Raithel \\
  BIFOLD\\
  Quality \& Usability Lab, TU Berlin\\
  DFKI GmbH\\ 
  Université Paris-Saclay, CNRS, LISN\\
  \texttt{raithel@tu-berlin.de} \\}

\begin{document}
\maketitle
\begin{abstract}
The NLI4CT task at SemEval-2024 emphasizes the development of robust models for Natural Language Inference on Clinical Trial Reports (CTRs) using large language models (LLMs).
This edition introduces interventions specifically targeting the numerical, vocabulary, and semantic aspects of CTRs. Our proposed system harnesses the capabilities of the state-of-the-art Mistral model \citep{jiang2023mistral}, complemented by an auxiliary model, to focus on the intricate input space of the NLI4CT dataset. Through the incorporation of numerical and acronym-based perturbations to the data, we train a robust system capable of handling both semantic-altering and numerical contradiction interventions. Our analysis on the dataset sheds light on the challenging sections of the CTRs for reasoning.
\end{abstract}

\section{Introduction}

Over the last decade, Natural Language Processing (NLP) has seen significant advancements, beginning with the introduction of word embeddings \citep{mikolov2013distributed}, followed by transformer architectures like BERT \citep{vaswani2017attention, devlin2018bert}, and specialized language models (LMs) such as BioBERT \citep{lee2020biobert} and PubMedBERT \citep{gu2021domain} tailored for the biomedical domain. The advent of large language models (LLMs) like GPT-3 \cite{brown2020language}, commonly known as Chat-GPT, has further pushed the boundaries of NLP, showcasing capabilities in diverse NLP tasks and even reasoning. However, LLMs adapt to shortcut learning easily instead of understanding the task at hand and resorting to shallow lexical heuristics for making a prediction \citep{tsuchiya2018performance, poliak2018hypothesis, naik2018stress}. Additionally, we have seen generative models like Chat-GPT hallucinating, making false claims, and struggling with providing factual information \citep{elazar2021measuring, wang2023survey}. Tackling these challenges is essential for ensuring the reliable deployment of large language models, particularly in critical fields like biomedicine, where the margin for error must be minimized.

The SemEval-2024 Task 2: \textit{Safe Biomedical Natural Language Inference for Clinical Trials} is focused on improving the understanding and evaluation methodologies for Large Language Models in clinical Natural Language Inference (NLI) \cite{jullien-etal-2024-semeval}.  This task targets aspects such as numerical and quantitative reasoning, domain-specific terminology, syntax, and semantics. It aims to analyze models' robustness, consistency, and faithfulness in reasoning within the clinical domain.

Our approach to this task involved leveraging instruction fine-tuned LLMs along with an auxiliary model that focuses on ``hard'' instances to develop a more resilient NLI system. ``Hard'' instances refer to those examples in the dataset where the model fails. Building on the methodology outlined by \citet{kanakarajan2023saama}, we assessed the zero-shot performance of various instruction-tuned LLMs to identify the most effective model. Upon selecting the best LLM, we introduced an auxiliary module during the fine-tuning process, which emphasized learning ``hard'' examples. Taking inspiration from \citet{korakakis2023improving}, who experimented with various configurations for the auxiliary module and highlighted its substantial impact on the final NLI system's performance, we explored various architectures for this auxiliary module. To improve the robustness of the system and address challenges related to numerical reasoning and domain-specific terminology, we introduced numerical and semantic perturbation to the NLI4CT dataset and trained our system on these. Our system ranked 11th in \textit{macro $F_1$ score}, 12th in \textit{Faithfulness}, and 19th \textit{Consistency} out of 31 participants. Our final system struggled when dealing with semantic-preserving interventions on the test data yet demonstrated strong performance on semantic-altering interventions.

\section{Background}

We now provide a description of the shared task, followed by a brief overview of the NLI4CT dataset. 
We then explore existing research, assessing their strengths and limitations while also drawing connections to our proposed method.

\subsection{Task and Dataset Description}

\begin{figure}[t]
    \centering
    \includegraphics[scale=0.38]{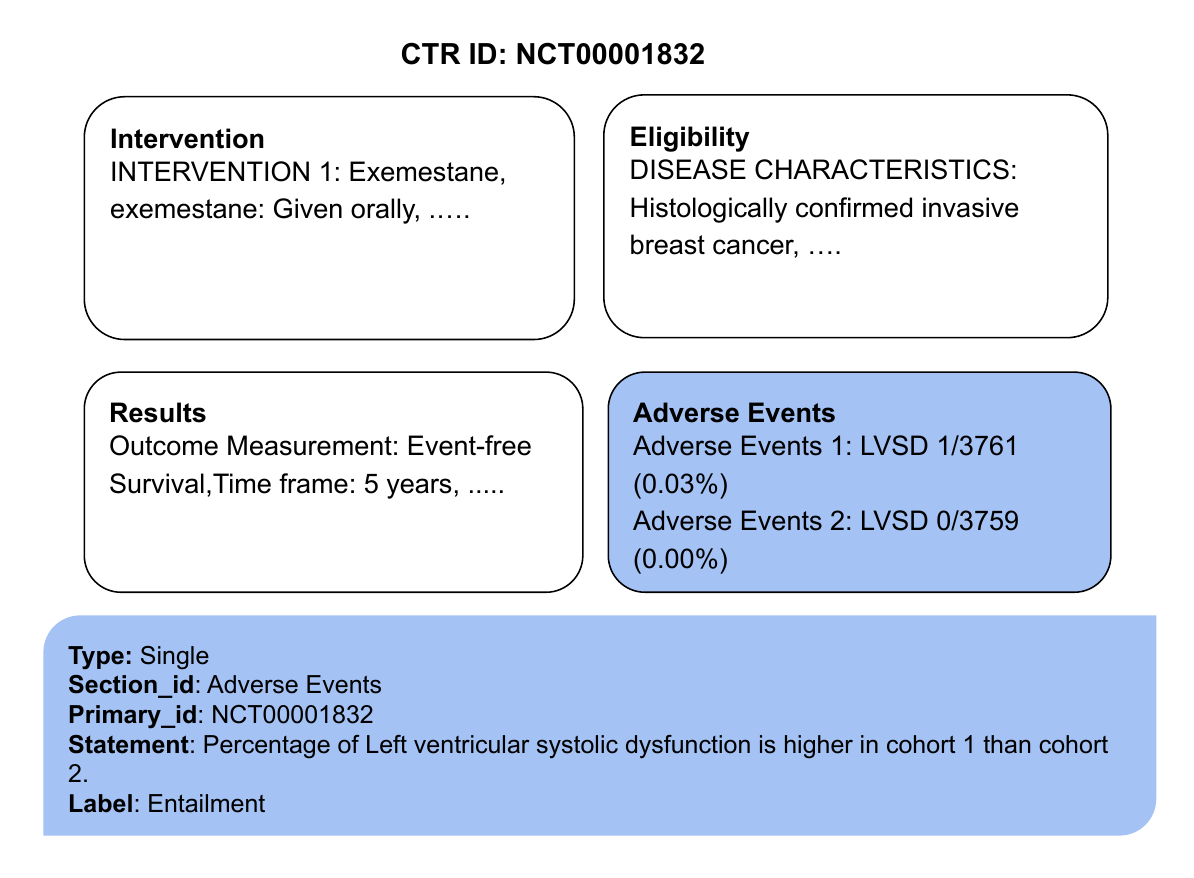}
    \caption{A sample instance from the NLI4CT dataset. 
    Each instance consists of four sections: \texttt{Intervention}, \texttt{Eligibility criteria}, \texttt{Results}, and \texttt{Adverse Events}. 
    The data are split into two types: \textit{single} (depicted) and \textit{comparison}. 
    In \textit{single}, one section of the CTR serves as the premise (in this case, \texttt{Adverse Events}). 
    A human-annotated hypothesis for this premise is given (Statement), which is then to be classified into either \textbf{entailment} or \textbf{contradiction}.}
    \label{fig:nli4ct_sample}
\end{figure}

This task is a continuation from SemEval-2023 Task 7 \cite{valentino2023semeval}, which introduced the NLI4CT dataset \cite{jullien2023nli4ct} derived from Clinical Trial Reports (CTRs) on breast cancer. The dataset contains 999 CTRs, each of which consists of four sections: \texttt{Eligibility Criteria}, a set of conditions for patients to be allowed to take part in the clinical trial; \texttt{Intervention}, information concerning the type, dosage, frequency, and duration of treatments being studied; \texttt{Results}, the number of participants in the trial, outcome measures, units, and the results; and \texttt{Adverse Events}, signs and symptoms observed in patients during the clinical trial. The dataset comprises two types of training instances: \textit{single} and \textit{comparison}. In the \textit{single} instances, one section of the CTR serves as the premise, while a corresponding human-annotated statement is presented as the hypothesis. On the other hand, in the \textit{comparison} instances, the same section of two CTRs is utilized, and the hypothesis typically involves a human-annotated comparative statement between the two CTRs. Each instance is labeled either \textbf{entailment} or \textbf{contradiction}, with an equal distribution of proportions between the two labels (more details in \Cref{app:dataset_stats}).
A sample instance for \textit{single} is shown in \Cref{fig:nli4ct_sample}.

\subsection{Related Works}

The NLI4CT dataset \citep{jullien2023nli4ct} was introduced in SemEval 2023 Task 7 \citep{valentino2023semeval}, where multiple submissions highlighted the aforementioned challenges associated with language models. The second-ranked team from SemEval 2023 Task 7, Saama AI Research \citep{kanakarajan2023saama}, initially evaluated an instruction-tuned LLM in a zero-shot setting. Subsequently, they fine-tuned the model using the best instruction with T5 \citep{raffel2020exploring} and Flan-T5-XXL \citep{chung2022scaling}. Motivated by their methodology and drawing inspiration from recent advancements, we employed instruction-tuned LLMs such as Llama \citep{touvron2023llama} and Mistral \citep{jiang2023mistral}, which represent state-of-the-art LLMs. Additionally, building upon the work of \citet{korakakis2023improving}, who introduced a learner-auxiliary model framework to enhance the robustness of NLI, we aimed to integrate this framework alongside the use of instruction-tuned LLMs in our approach.

The challenge of word distribution shift from the general domain to the biomedical domain has posed a significant obstacle to the effectiveness of NLP methods applied to the biomedical field. 

The prevalence of aliases and acronyms in biomedical text prompted \citet{jin2019deep} to propose a model that automatically collects context for abbreviations from PubMed abstracts and employs a BiLSTM classifier for abbreviation expansion. 
Additionally, \citet{grossman2021deep} presented a Medical Abbreviation and Acronym Meta-Inventory\footnote{\url{https://github.com/lisavirginia/clinical-abbreviations}}, constituting a comprehensive database of medical abbreviations encompassing 104,057 entries, each linked to 170,426 corresponding senses. We leveraged this Meta-Inventory to incorporate acronym-based perturbations into the NLI4CT dataset. Additionally, we also incorporated a pre-finetuning phase into our approach by fine-tuning on the MedNLI\footnote{\url{https://physionet.org/content/mednli/1.0.0/}} dataset \citep{shivade2019mednli}. This step aims to familiarize the model with clinical data. 
\section{System Overview}

In light of recent advancements in large language models and drawing insights from the results of the SemEval 2023 Task 7 \cite{valentino2023semeval}, we implemented the approach outlined in the work of \citet{kanakarajan2023saama}. 
Our approach involved evaluating state-of-the-art LLMs, including Mistral \cite{jiang2023mistral}, Llama \cite{touvron2023llama}, and Lemma \cite{azerbayev2023llemma}, alongside their variants. 
We experimented with different instructions for each model and subsequently compared their zero-shot performance based on their respective best-performing instruction (see final instruction template in \Cref{app:instruction_template}). Mistral emerged as the top-performing model among all others evaluated with the highest $F_1$ score (0.69). Furthermore, we implemented the MinMax algorithm \cite{korakakis2023improving} by adding an auxiliary model alongside the Mistral model to create a more robust system. This auxiliary model is designed to amplify the loss incurred in input spaces where the Mistral model encounters difficulties, effectively directing its focus towards areas of higher loss. To further boost the performance of the system, we pre-finetuned using MedNLI dataset. Additionally, we conducted an error analysis to identify easy and difficult instances in the train set to provide a basis for further research.

\begin{table*}[t]
\centering
\small
\begin{tabular}{@{}lcccc@{}}
\toprule
\textbf{Model} & \textbf{Dev $F_1$} & \textbf{Test $F_1$} & \textbf{Consistency} & \textbf{Faithfulness} \\
\midrule
NLI4CT-FT & 0.69 & 0.74 & 0.68 & 0.75 \\
NLI4CT-FT-ACR & 0.73 & 0.76 & 0.67 & 0.71 \\
MEDNLI-FT-NLI4CT & 0.75 & 0.75 & 0.68 & 0.78 \\
MEDNLI-FT-NLI4CT-ACR-NUM & 0.75 & 0.74 & 0.68 & 0.78 \\
MEDNLI-NLI4CT-FT-ACR & 0.74 & 0.75 & 0.67 & 0.74 \\
MEDNLI-NLI4CT-FT-NUM & 0.74 & 0.73 & 0.69 & 0.79 \\
MEDNLI-NLI4CT-FT-ACR-NUM & 0.75 & \textbf{0.76} & \textbf{0.70} & 0.75 \\
MINMAX-MEDNLI-FT-NLI4CT & 0.75 & 0.75 & 0.68 & \textbf{0.82} \\
MINMAX-MEDNLI-FT-NLI4CT-BC & \textbf{0.77} & 0.75 & 0.68 & 0.78 \\
MINMAX-MEDNLI-NLI4CT-FT-ACR-NUM-BC & 0.74 & 0.74 & 0.68 & 0.75 \\
\bottomrule
\end{tabular}%
\caption{Final results on the NLI4CT dataset. \textbf{Dev $F_1$} and \textbf{Test $F_1$} represent the \textit{macro} $F_1$ score on the development  and test set, respectively. \textbf{Consistency} measures the ability to predict same labels for \textit{semantic preserving} interventions and \textbf{Faithfulness} measures the ability to correctly change the labels for \textit{semantic altering} interventions. Both \textit{Consistency} and \textit{Faithfulness} results are on the test set.}
\label{table:metrics}
\end{table*}

\section{Experimental Setup}

Training an LLM can be both costly and resource-intensive. However, recent advancements in methodologies, such as Parameter-Efficient Fine-Tuning (PEFT), have emerged to reduce the computational cost of fine-tuning \cite{peft}.
For fine-tuning the Mistral model, we employ a PEFT method known as Low-Rank Adaption (LoRA, \citet{hu2021lora}). We adopted a similar approach for implementing the auxiliary model as described by \citet{korakakis2023improving}.  We experimented with the parameters of the system to obtain an optimal architecture, details of which can be found in \Cref{app:experiment_details}. 

\subsection{Data Perturbation}

Utilizing the Meta-Inventory of \citet{grossman2021deep}, we extracted the short forms from 358 NLI4CT hypotheses, resolving them to their corresponding long forms based on the cosine similarity. 
This resulted in 352 perturbed instances with consistent labels. Additionally, 181 negative instances were generated by selecting the least similar long forms, resulting in a total of 533 new instances for the acronym-based perturbation. For numerical perturbations, we employed an English Named Entity Recognition model \cite{raza2022large} trained on Maccrobat to extract 27 unique biomedical entities from hypotheses. We perturbed numerical values and introduced semantic alterations that generated 355 new instances with labels flipped. For more details, see \Cref{app:data_pert_description}.

\subsection{Fine-tuning Strategies}
We performed various experiments involving different combinations of fine-tuning methodologies. Initially, we fine-tuned only the Mistral model (\textit{NLI4CT-FT}) on NLI4CT without incorporating the auxiliary model. An extension of this initial setup involved N-step fine-tuning, where, for example, in a two-step fine-tuning approach, we first fine-tuned the model with the MedNLI dataset and subsequently fine-tuned it further with the NLI4CT dataset (\textit{MEDNLI-FT-NLI4CT}). We proceeded to add more steps by fine-tuning on perturbated datasets, such as the acronym-perturbed dataset (\textit{MEDNLI-NLI4CT-FT-ACR}) or the numerically-perturbed dataset (\textit{MEDNLI-NLI4CT-FT-NUM}). The MinMax algorithm requires that the base model be trained for a few epochs or steps. We utilized the best-performing models from previous N-step experiments to adapt this strategy effectively. 

This way, we already have a model that is trained on the dataset and add the auxiliary model to enhance the robustness of the whole system. Details of all the models that we fine-tuned with different strategies can be found \Cref{app:model_description}.

\subsection{Evaluation Strategies}
In our initial experiments, we observed that Mistral 7B exhibited superior performance compared to Mistral Instruct 7B post-fine-tuning. Consequently, we primarily trained most models using Mistral 7B. However, during the evaluation phase, we attached the PEFT fine-tuned adapter with both Mistral and Mistral Instruct 7B to compare their results. To stabilize the model's generation behavior, we conduct evaluations on the development set five times and select the label predicted most frequently across these five runs. Similarly, for test data, we perform three runs.

\section{Results}

During both the fine-tuning and evaluation phases, we observed improvements in the model trained with the MinMax algorithm compared to other models. From \Cref{table:metrics}, we can see the model (\textit{MINMAX-MEDNLI-FT-NLI4CT-BC}) trained with the MinMax algorithm achieved the highest $F_1$ score on the dev set. When comparing the base model (\textit{MEDNLI-FT-NLI4CT}) with the MinMax model (\textit{MINMAX-MEDNLI-FT-NLI4CT}), we noted a slight improvement in \textit{Consistency} and a significant improvement in \textit{Faithfulness}. Although the $F_1$ score did not exhibit improvement, the enhancements in the other metrics indicate that the MinMax algorithm contributed to the development of a more robust system and was able to handle the semantically altering intervention much better. Regarding the models trained with perturbed data, we observed a negative effect on the overall performance of the MinMax-trained model (\textit{MINMAX-MEDNLI-NLI4CT-FT-ACR-NUM-BC}) compared to the base model (\textit{MEDNLI-NLI4CT-FT-ACR-NUM}). For our final submission to the leaderboard, we submitted the MinMax model (\textit{MINMAX-MEDNLI-FT-NLI4CT}), which ranked 11th in \textit{macro $F_1$ score}, 12th in \textit{Faithfulness}, and 19th in \textit{Consistency}.

\subsection{Impact of Data Perturbation}

To assess the impact of acronym-based perturbed data, we initially trained the model using the original NLI4CT dataset and subsequently fine-tuned it with the acronym-based data. Evaluation of both models was conducted on the test data, which comprise the following intervention types introduced by the task's organizers: Control, Contrast, Paraphrase, Contradiction, Numerical Contradiction, Numerical Paraphrase, and Definitions. For accessing a model trained on acronym-based perturbed data, we look at the metrics for the intervention types Paraphrase (\textit{Para}) and Definitions. \Cref{tab:pert_data_metrics} in \Cref{app:perturbations} shows that acronym perturbation notably enhanced results for the intervention-type Definitions. Similarly, with numerical-based perturbation, we look at metrics for intervention-type Numerical Paraphrase (\textit{Num\_Para}) and Numerical Contradiction (\textit{Num\_Cont}). While no changes were observed in the results of semantic-altering interventions, there was some improvement noted in semantic-preserving interventions. Lastly, we investigated the combined impact of both perturbations and their influence on all four interventions. Overall, we observed that combined fine-tuning improved the Definition and Paraphrase intervention type more than the Numerical intervention types. However, there was a negative impact observed on semantic-altering numerical interventions.

\subsection{Performance across Interventions and Sections}
\label{sec:res_int_sec}

\Cref{tab:intervention_section_based_results} in \Cref{app:intervention_results} presents the results of the test data across various interventions and sections. The \texttt{Adverse Events} section exhibits the highest $F_1$ score at 0.73, whereas the \texttt{Eligibility} section demonstrates the lowest score at 0.66. In terms of interventions, Numerical\_contradiction achieves the highest score at 0.93, while Definition attains the lowest at 0.63. 
Among the interventions featuring both labels, Paraphrase achieves the highest performance with an $F_1$ score of 0.72. Moreover, it is the only intervention type that achieves a higher score for \textbf{entailment}. Conversely, all other sections and interventions exhibit better performance for the \textbf{contradiction} label.

\section{Error Analysis}

To understand the model's behavior across different sections, interventions, and labels/relations, we examined the dataset.

\subsection{Dataset Difficulty}

\begin{figure}[h]
    \centering
    \includegraphics[scale=0.4]{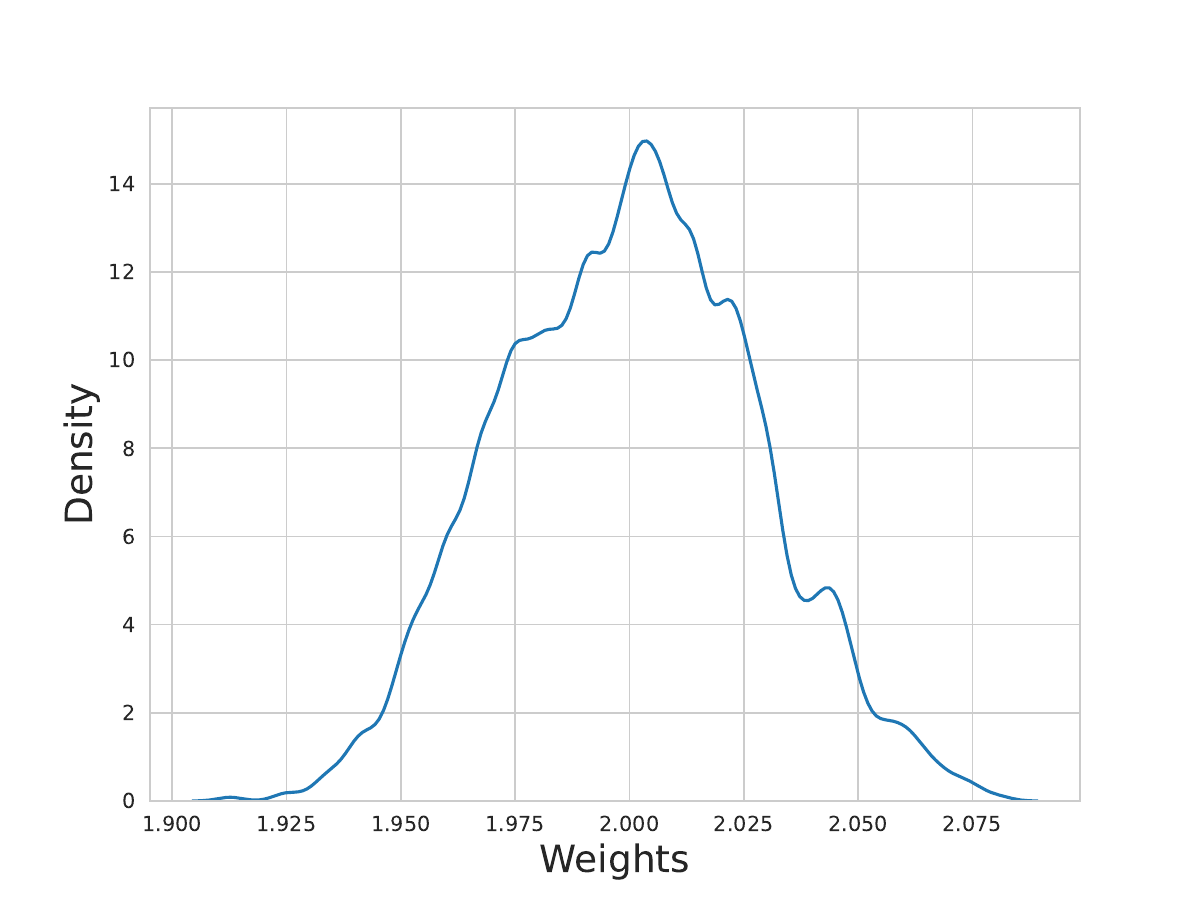}
    \caption{Weight distribution of NLI4CT data instances generated by the auxiliary model after 3 epochs of training. Lower weights correspond to easy examples, and higher weights correspond to hard examples.}
    \label{fig:mm_wgts}
\end{figure}

One application of the MinMax algorithm is its capability to classify data points into hard and easy examples. \Cref{fig:mm_wgts} illustrates the weight distribution of data instances from the auxiliary model after three epochs of training. Data instances with higher weights represent hard examples, where the model incurs a high loss, while instances with lower weights denote easy examples.

\begin{figure}[t]
    \centering
    \includegraphics[scale=0.38]{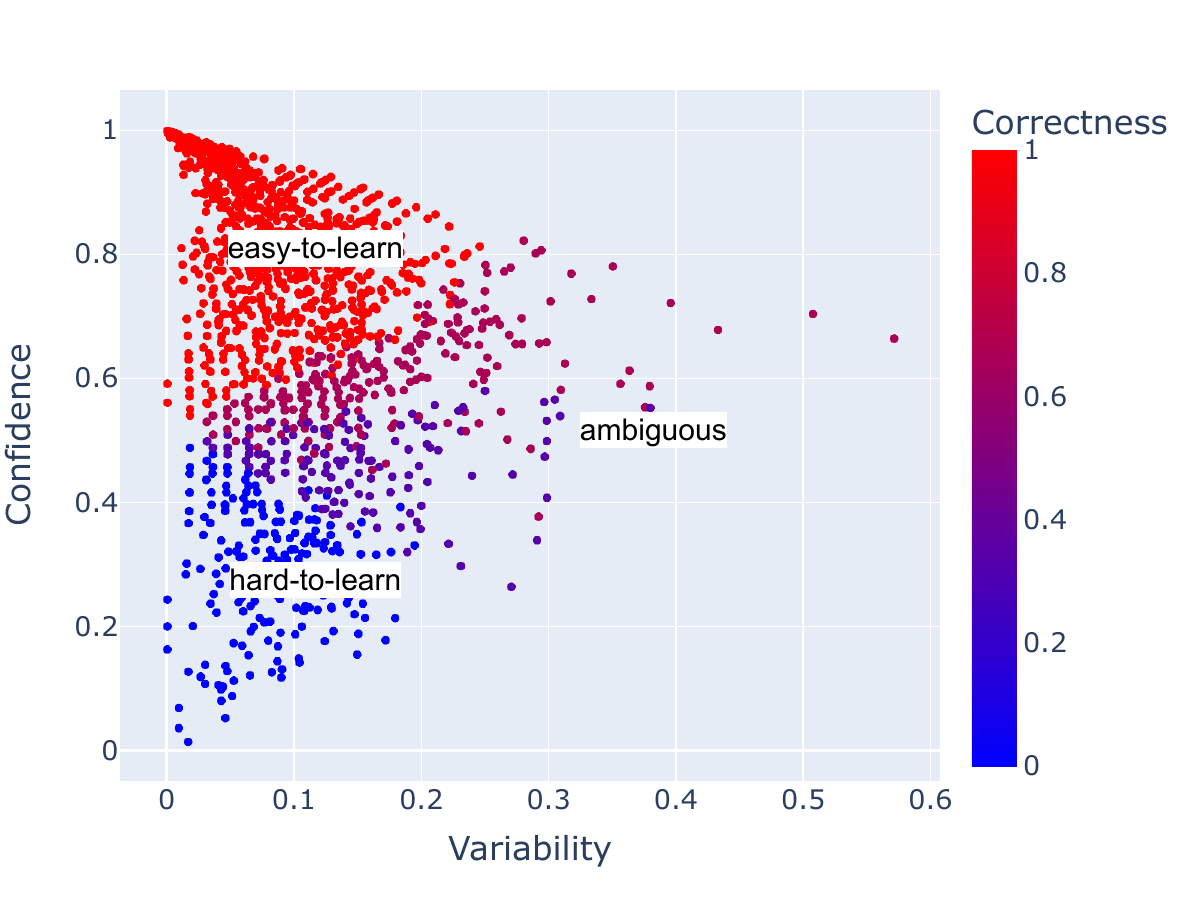}
    \caption{Data map for the NLI4CT dataset following \cite{swayamdipta2020dataset}.}
    \label{fig:dc_map}
\end{figure}

Following the data cartography procedure outlined in \citet{swayamdipta2020dataset}, we replicated their method using the best MinMax model trained for three epochs. We collected probability values for the gold label on each epoch and calculated confidence, variability, and correctness values. In  \Cref{fig:dc_map}, the upper region with red data points represents easy-to-learn instances, while the bottom region with blue data points represents hard-to-learn examples. Data points with high variability are depicted as ambiguous examples.

\subsection{Analysis of Easy and Hard Samples}

We conducted a comparison between the two dataset difficulty methodologies by extracting easy and hard examples from both strategies. We found 322 instances common to both strategies as easy examples or easy-to-learn instances. As for hard examples or hard-to-learn examples, there were 96 instances common to both. We performed a three-level analysis using these instances, especially the hardest ones, to understand the in-depth dataset difficulty and the model’s behavior. 
\begin{figure*}[t]
    \centering
    \includegraphics[scale=0.5]{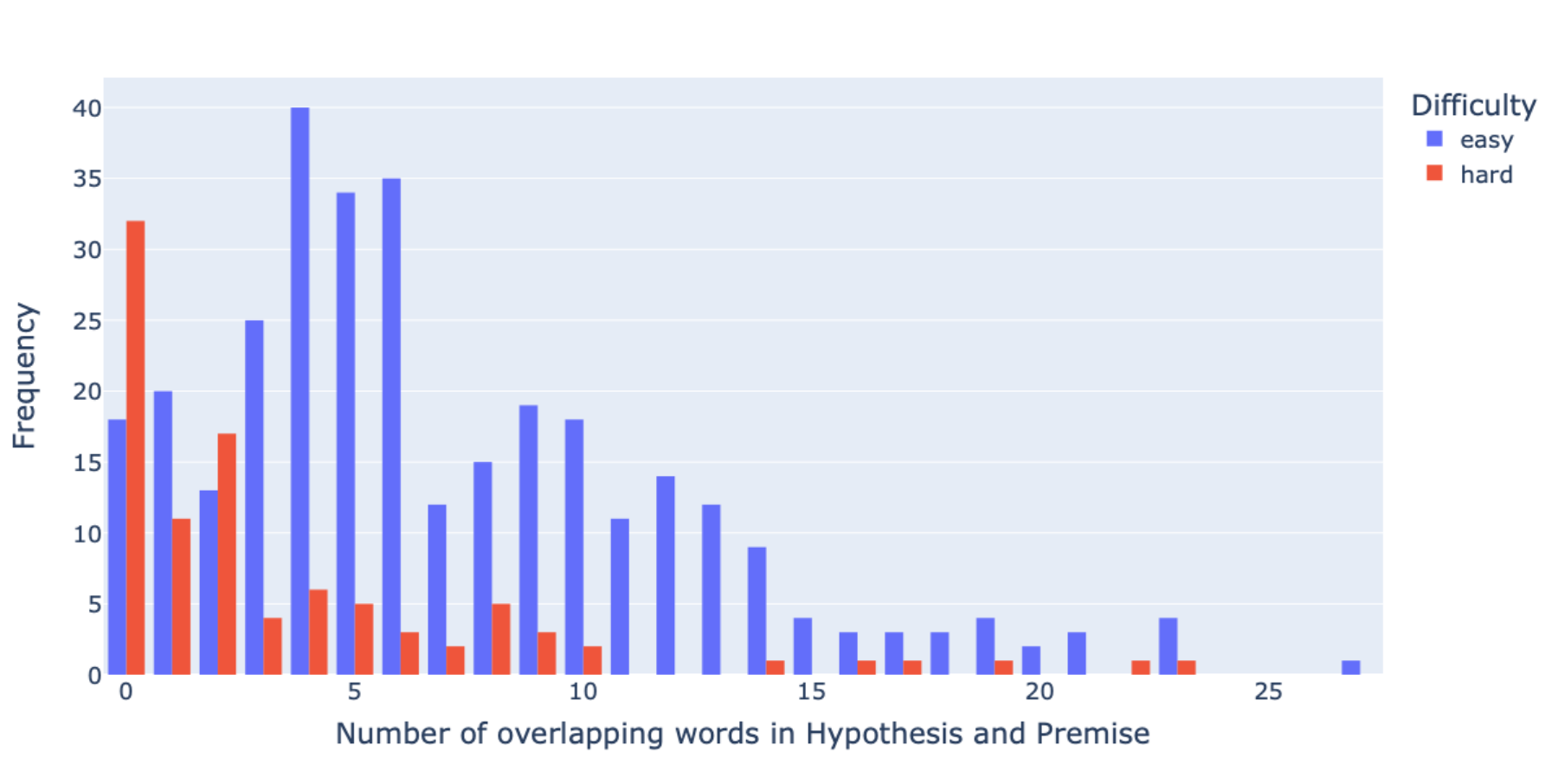}
    \centering
    \caption{Word overlap between the hypothesis and the premise in the easy and the hard examples. }
    \label{fig:word_overlap}
\end{figure*}

First, we looked at the structural level of the dataset concerning these instances and found that instances focusing on the \texttt{Eligibility} section were identified as the most easy-to-learn for the model, whereas those targeting \texttt{Adverse Events} proved challenging. Additionally, learning the \textbf{contradiction} relation was more difficult than \textbf{entailment} (see \Cref{tab:data_difficulty_stats}). Next, we compared the word overlap between the premise and hypothesis of the easy and hard examples. We found that the word overlap was higher in the easy examples compared to the hard examples. Furthermore, the easy examples exhibited a higher frequency of \textbf{entailment} relations, suggesting that the model might have established a correlation between word overlap and \textbf{entailment} relations (see \Cref{fig:word_overlap}). One potential solution to mitigate this issue could involve perturbing the instances with high word overlap by introducing synonyms into the dataset. 

Combining observations from these analyses provides some interesting insights. As previously discussed in \ref{sec:res_int_sec}, the results from the test data reveal that the \texttt{Eligibility} section obtained the lowest $F_1$ score, while \texttt{Adverse Events} performed the best. Given that the instances of the \texttt{Eligibility} section in the training set were easy to learn, it is plausible that the model did not learn many features from this section. Conversely, as the instances of \texttt{Adverse Events} were more challenging to learn, the model likely attempted to extract more features from this section. A similar rationale can also be applied to the \textbf{entailment} and \textbf{contradiction} relation. However, another factor contributing to the higher scores on the \textbf{contradiction} relations in the test data could be the greater number of the true \textbf{contradiction} relations.

Finally, we manually analyzed the ten most difficult examples. We discovered that the predominant error made by the model involved the confusion between the cohorts and the trials. Specifically, the instances that involve a comparison between two trials, each comprising two cohorts, often led the model to misinterpret the second cohort of the first trial as the secondary trial. Overall, the model struggled with numerical reasoning, particularly in scenarios involving numerous variables that require calculations. More details on the analysis of dataset difficulty can be found in \Cref{app:dataset_difficulty}.

\section{Conclusion}

In this study, we introduced a large language model-based system designed to address the natural language inference task through text generation. Our approach prioritized model robustness, which was achieved by incorporating an auxiliary model that directs the LLM to focus on challenging instances in the input space. Moreover, we enhanced the system's robustness against adversarial samples by introducing numerical and semantic perturbations to the NLI4CT dataset during training. Our findings revealed the system's superior robustness against semantic-altering interventions compared to semantic-preserving ones. Additionally, through dataset analysis, we identified instances targeting the \texttt{Eligibility} section in Clinical Trial Reports as the the easiest to learn but more challenging to predict accurately. Conversely, the \texttt{Adverse Events} section posed greater difficulty in learning but was relatively easier to predict accurately. These findings offer valuable insights for future research on improving the robustness by focusing more on challenging sections of CTRs.

\section*{Acknowledgments}

Our work was funded by the Deutsche Forschungsgemeinschaft (German Research Foundation) DFG-442445488 under the trilateral ANR-DFG-JST AI research project KEEPHA.
Furthermore, we gratefully acknowledge funding from the German Federal Ministry of Education and Research under the grant BIFOLD24B.

\section*{Limitations and Ethical Considerations}

We do not rule out the possible risk of sensitive content in the data. 
Furthermore, the Mistral-based models in our experiments, which were pre-trained on a wide variety of source data, might have inherited biases from these pretraining corpora.
We further acknowledge that prompts used to generate responses with Mistral models might result in different responses when the prompts are slightly modified or set up differently.

\bibliography{main}

\clearpage

\appendix

\section{Appendix}
\label{sec:appendix}

\subsection{Dataset Statistics}\label{app:dataset_stats}
We highlight the basic statistics of the NLI4CT dataset in \Cref{tab:dataset_stats}.

\begin{table*}[t]
  \small\centering
   \resizebox{\textwidth}{!}{
  \begin{tabular}{lccccccccc}

    \toprule

    {\bfseries Data} & {\bfseries No of Samples} & \multicolumn{2}{|c|}{{\bfseries Type}} & \multicolumn{4}{|c|}{{\bfseries Section}} & \multicolumn{2}{|c|}{{\bfseries Label}} \\
    \midrule\midrule[.1em]
     &  & & Count & \texttt{Intervention} & \texttt{Eligibility} & \texttt{Adverse Events} & \texttt{Results} & contradiction & entailment \\
    \midrule\midrule[.1em]

    \multirow{2}{*}{Train}
    & \multirow{2}{*}{1700} & \textit{single} & 1035 & 155  & 317 & 309 & 254  & 502 & 533 \\
    &  & \textit{comparison} & 665 & 241  & 169 & 187 & 68  & 348 & 317 \\

    \midrule[.1em] 
    
    \multirow{2}{*}{Dev}
    & \multirow{2}{*}{200} & \textit{single} & 140 & 26  & 44 & 32 & 38  & 70 & 70 \\
    &  & \textit{comparison} & 60 & 10  & 12 & 20 & 18  & 30 & 30 \\

    \midrule[.1em] 
    
    \multirow{2}{*}{Test}
    & \multirow{2}{*}{5500} & \textit{single} & 2553 & 784  & 468 & 523 & 778  & 1703 & 850 \\
    &  & \textit{comparison} & 2947 & 758  & 951 & 781 & 457  & 1956 & 991 \\

    \bottomrule
  \end{tabular}%
   }
  \centering
\caption{NLI4CT statistics.}
\label{tab:dataset_stats}
\end{table*}

\subsection{Descriptions of the Fine-tuned Models}
\label{app:model_description}

We implemented various fine-tuning strategies across multiple models. Below, we provide descriptions for each of these models:
\begin{itemize}
    \item \textbf{NLI4CT-FT}: Mistral-7B model fine-tuned on the NLI4CT dataset.
    \item \textbf{MEDNLI-FT}: Mistral-7B model fine-tuned on the MEDNLI dataset.
    \item \textbf{NLI4CT-FT-ACR}: NLI4CT-FT model fine-tuned on acronym based perturbations. 
    \item \textbf{MEDNLI-FT-NLI4CT}: MEDNLI-FT fine-tuned on the NLI4CT dataset.
    \item \textbf{MEDNLI-FT-NLI4CT-ACR-NUM}: MEDNLI-FT fine-tuned simultaneously on the NLI4CT dataset, acronym and numerical perturbations.
    \item \textbf{MEDNLI-NLI4CT-FT-ACR}: MEDNLI-FT-NLI4CT model fine-tuned on acronym perturbations.
    \item \textbf{MEDNLI-NLI4CT-FT-NUM}: MEDNLI-FT-NLI4CT model fine-tuned on numerical perturbations.
    \item \textbf{MEDNLI-NLI4CT-FT-ACR-NUM}: MEDNLI-FT-NLI4CT model fine-tuned on acronym and numerical perturbations simultaneously.
    \item \textbf{MINMAX-MEDNLI-FT-NLI4CT}: MEDNLI-FT fine-tuned using Mistral-7B and the auxiliary model on the NLI4CT dataset.
    \item \textbf{MINMAX-MEDNLI-FT-NLI4CT-BC}: MEDNLI-FT fine-tuned using Mistral-7B and the auxiliary model on the NLI4CT dataset using the best configuration obtained from hyperparameter tuning.
    \item \textbf{MINMAX-MEDNLI-NLI4CT-FT-ACR-NUM-BC}: MINMAX-MEDNLI-FT-NLI4CT fine-tuned using Mistral-7B and the auxiliary model on acronym and numerical perturbations simultaneously using the best configuration obtained from hyperparameter tuning.
    
\end{itemize}

\subsection{Final Instruction Template}
\label{app:instruction_template}

\begin{figure}[t]
    \centering
    \includegraphics[scale=0.3]{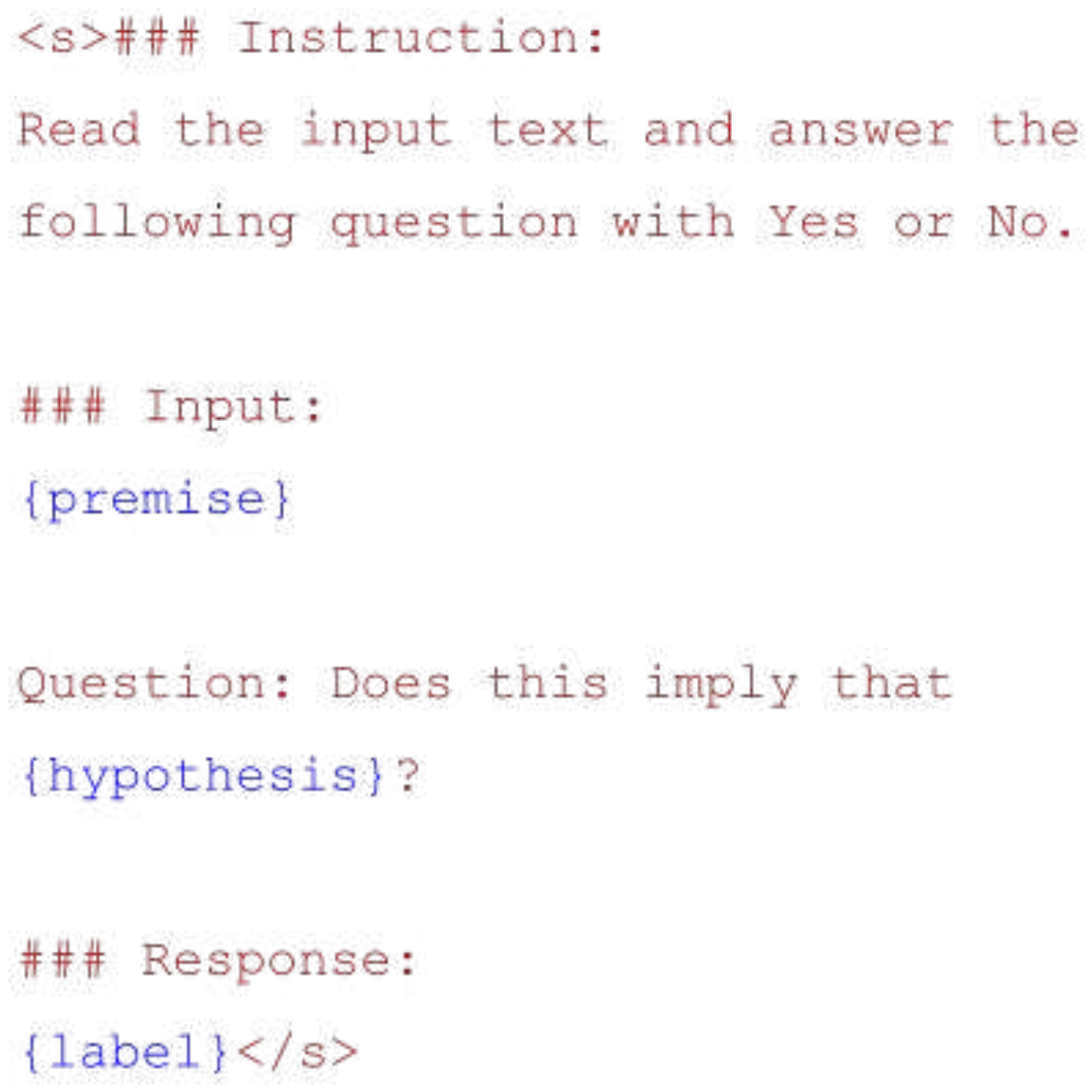}
    \caption{Final design for prompting.}
    \label{fig:prompt}
\end{figure}

After running experiments with different prompt formats, we finalized the template as shown in \Cref{fig:prompt}.  
Instead of directly tackling the NLI task, we frame it as a text generation problem. We begin by giving general instructions, which describe the task to be performed. The next two sections of the prompt consist of the premise, providing context for the task, and the hypothesis presented as a question. The model is then trained to generate either ``Yes'' for an \textbf{entailment} relationship between the premise and hypothesis or ``No'' for a \textbf{contradiction}. While fine-tuning our model with the MedNLI dataset, we only utilized entailment and contradiction instances, excluding those labeled as neutral, to ensure consistency with the NLI4CT dataset.

\subsection{Data Perturbation Details}
\label{app:data_pert_description}
In \Cref{tab:pert_data_stats}, we show full statistics of data perturbation on NLI4CT dataset. In the following section, we describe the data perturbation methodology. 

\subsubsection{Acronym Based Perturbations}

\begin{table*}[h]
  \small\centering
  \resizebox{\textwidth}{!}{
  \begin{tabular}{lccccccccc}

    \toprule

    {\bfseries Data} & {\bfseries No of Samples} & \multicolumn{2}{|c|}{{\bfseries Type}} & \multicolumn{4}{|c|}{{\bfseries Section}} & \multicolumn{2}{|c|}{{\bfseries Label}} \\
    \midrule\midrule[.1em]
     &  & & Count & \ttfamily{Intervention} & \ttfamily{Eligibility} & \ttfamily{Adverse Events} & \ttfamily{Results} & contradiction & entailment \\
    \midrule\midrule[.1em]

    \multirow{2}{*}{ACR}
    & \multirow{2}{*}{533} & \textit{single} & 357 & 67  & 103 & 20 & 167  & 178 & 179 \\
    &  & \textit{comparison} & 176 & 46  & 70 & 53 & 7  & 93 & 83 \\

    \midrule[.1em] 
    
    \multirow{2}{*}{NUM}
    & \multirow{2}{*}{355} & \textit{Single} & 268 & 51  & 67 & 39 & 111  & 267 & 1 \\
    &  & \textit{comparison} & 87 & 24  & 25 & 31 & 7  & 86 & 1 \\

    \bottomrule
  \end{tabular}%
}
  \centering
\caption{Statistics for Acronym the (ACR) and Numerical (NUM) based perturbed dataset across different sections, labels, and instance types.}
\label{tab:pert_data_stats}
\end{table*}

We utilized a Medical Abbreviation and Acronym Meta-Inventory \cite{grossman2021deep} containing short forms (SF) and corresponding long forms (LF) commonly used in the biomedical domain. With regular expressions, we extracted short forms present in the hypotheses of the NLI4CT dataset, resulting in 358 hypotheses. Given that the meta-inventory often includes multiple long forms for a single short form we computed the cosine similarity between the short forms in the hypotheses and their corresponding long forms in the meta-inventory. For each unique short form identified in the hypotheses, we determined the most similar long form and manually verified its correctness within the context of the hypothesis. Subsequently, we resolved the short forms in the format: `SF (LF)'. This process yielded 352 perturbed instances with consistent inference labels. Such perturbations are intended to assist models in avoiding potential confusion by ensuring that short forms are resolved, even when their corresponding long forms are present in the premise. Likewise, for each unique short form, we identified the least similar long form and generated a negative instance following the same format as before. This process resulted in approximately 181 new negative instances, where labels were flipped. Consequently, when combining both acronym-based perturbations, we created a total of 533 new instances.

\subsubsection{Numerical Perturbation}

The Math Word Problem (MWP) task has been introduced in NLP to enhance models' numerical reasoning capabilities \cite{he2023solving, yao2023solving}. Within our dataset, numerous instances involve comparisons of numerical entities, which inherently qualify as MWPs. To augment these instances, we introduce noise to the numerical entities in various forms. Utilizing an English Named Entity Recognition model \cite{raza2022large} trained on Maccrobat, specifically tailored for biomedical entities (107 entities), we extracted 27 unique entities from the hypotheses. Our focus was on identifying entities that can alter the hypothesis's meaning concerning numerical reasoning, such as Age, Dosage, Lab\_value, Duration, and Date. For numerical values associated with these entities, we applied basic mathematical operations like addition or subtraction. Additionally, words comparing these numerical entities were replaced with their opposites; for example, `lower' was substituted with `higher', and `more than a week' was replaced with `less than a week', and so forth. This process resulted in a total of 355 new perturbed instances, each with its label flipped.

\subsubsection{Data Perturbations Results}
\label{app:perturbations}
\Cref{tab:pert_data_metrics} presents results for various interventions introduced in test data. \textit{Base} in the table refers to \textit{MEDNLI-FT-NLI4CT}.
\begin{table*}[h]
  \small
  \centering
  \begin{tabular}{lcccccccc}

    \toprule

    {\bfseries Model} & \multicolumn{3}{|c|}{{\bfseries $F_1$}} & \multicolumn{1}{|c|}{{\bfseries Faithfulness}} & \multicolumn{4}{|c|}{{\bfseries Consistency}} \\
    \midrule\midrule[.1em]
     & Definitions & Para & Num\_Para & Num\_Cont & Definitions & Para & Num\_Para & Num\_Cont \\
    \midrule\midrule[.1em]
    
    Base & 0.42 & 0.72 & 0.54 & 0.88 & 0.59 & 0.72 & 0.68 & 0.90\\
    Base + ACR & 0.49 & 0.73 & \textbf{0.59} & 0.82 & 0.61 & 0.71 & 0.68 & 0.88\\
    Base + NUM & 0.46 & 0.73 & 0.56 & \textbf{0.88} & 0.60 & 0.72 & 0.68 & \textbf{0.91}\\
    Base + ACR + NUM & \textbf{0.58} & \textbf{0.73} & 0.58 & 0.83 & \textbf{0.64} & \textbf{0.72} & \textbf{0.68} & 0.90\\
    MinMax + ACR + NUM & 0.51 & 0.73 & 0.56 & 0.83 & 0.62 & 0.71 & 0.68 & 0.88\\

    \bottomrule
  \end{tabular}%
  \centering
\caption{Acronym (ACR) and Numerical (NUM) perturbed dataset results}
\label{tab:pert_data_metrics}
\end{table*}

\subsection{Model and Experiment Details}
\label{app:modeling_details}
We provide information regarding the models, the minmax algorithm, and experiments.
\subsubsection{Mistral 7B and Mistral Instruct 7B}
\label{app:mistral_description}

Mistral 7B, as the name suggests, has 7 billion parameters and stands out as a language model engineered for exceptional performance and efficiency. Central to its architecture are the grouped-query attention \cite{ainslie2023gqa} and sliding window attention mechanisms \cite{child2019sparsetransformer, beltagy2020longformer}. Mistral models demonstrate remarkable adaptability and consistently outperform counterparts like Llama-13B. Moreover, the ease with which Mistral can be fine-tuned is evidenced by the Mistral Instruct 7B version, which is fine-tuned on publicly available instruction datasets and achieves a significant performance boost over the base version. Utilizing the capabilities of Mistral models, we fine-tuned both versions of the models on NLI4CT through a series of experiments aimed at determining the optimal version for final system development. 
Details of the Mistral models are shown in \Cref{table:tok_len_stats}.

\subsubsection{Low Rank Adaption}
\label{app:lora_description}
LoRA operates by freezing the weights of the pre-trained model and introducing trainable rank decomposition matrices into each layer of the Transformer architecture. This strategy significantly reduces the number of trainable parameters for downstream tasks, leading to lower memory usage and accelerated fine-tuning speed. 
We utilize the HuggingFace implementation of PEFT, which incorporates LoRA configurations to initialize LoRA-based fine-tuning of the Mistral model. By applying LoRA, we were able to reduce the number of training parameters from 3,837,112,320 to 85,041,152 (2.22\% of the total), which are subsequently optimized using the AdamW optimizer.

\subsubsection{MinMax Algorithm}
\label{app:minmax_description}
Beyond solely relying on the Mistral model, we introduced an auxiliary model into the fine-tuning process following the implementation of the MinMax algorithm introduced by \citet{korakakis2023improving} to enhance the model's robustness in NLI training. This auxiliary model is designed to amplify the loss incurred in input spaces where the Mistral model encounters difficulties, effectively directing its focus towards areas of higher loss. 
The objective function for training is defined as:
\[
J(\theta, \phi) = \min_{\theta} \max_{\phi} \frac{1}{n} \sum_{i=1}^{n} g_{\phi}(x_i, y_i) \cdot \mathcal{L}(f_{\theta}(x_i), y_i)
\]

Here, \( \theta \) denotes the mistral model parameters while \( \phi \) denotes the auxiliary parameters that are optimized using standard optimization methods. \( \mathcal{L}(f_{\theta}(x_i), y_i) \) is the cross entropy loss and \( g_{\phi}(x_i, y_i) \) generates weights for each instance in the range (0,1).

\subsubsection{Experiment Details}
\label{app:experiment_details}
Here we provide the parameters used in our experiments for both the base and auxiliary models. For the base Mistral model, we used a LoRA configuration with the following parameters: 

\begin{verbatim}
rank: 32
lora_alpha: 64
target_modules: [ q_proj, k_proj, v_proj,
o_proj, gate_proj, up_proj, down_proj,
lm_head ],
lora_dropout: 0.05
\end{verbatim}

Parameters for fine-tuning mistral and auxiliary models are as follows:
\begin{verbatim}
Mistral:
    learning_rate: 3.3e-5
    batch_size: 4
    number_of_epoch: 1
    max_steps: 1000
Auxiliary:
    learning_rate: 5.8e-3
    hidden_size_1: 1024
    hidden_size_2: 64
\end{verbatim}

Further system training and hyperparameter tuning details can be found at \url{https://github.com/Bhuvanesh-Verma/RobustLLM}

\subsection{Results of Best Model with respect to Interventions and Sections}\label{app:intervention_results}

We examined the results on test data across various sections and interventions. \Cref{tab:intervention_section_based_results} indicates that the \texttt{Adverse Event} section and \textit{Numerical Contradiction} interventions yield the best performance. 

\begin{table*}[h]
  \small\centering
  \resizebox{\textwidth}{!}{
  \begin{tabular}{lccccc}

    \toprule

    & {\bfseries Type} & {\bfseries No of Samples} & \multicolumn{3}{|c|}{{\bfseries $F_1$ Score}} \\
    \midrule\midrule[.1em]
    & & & entailment & contradiction & macro avg \\
    \midrule\midrule[.1em]

    \multirow{4}{*}{Section}
    & {\ttfamily Intervention} & 1542 & 0.58 (512)  & 0.75 (1030) & 0.67 \\
    & {\ttfamily Eligibility} & 1419 & 0.58 (485)  & 0.73 (934) & 0.66 \\
    & {\ttfamily Results} & 1235 & 0.58 (405)  & 0.80 (830) & 0.69 \\
    & {\ttfamily Adverse Events} & 1304 & 0.65 (439)  & 0.81 (865) & 0.73 \\

    \midrule[.1em] 
    
    \multirow{4}{*}{Intervention}
    &  Contradiction & 1500   & 0 (0)& 0.84 (1500) & 0 \\
    & Numerical\_contradiction & 276  & 0 (0) & 0.93 (276) & 0 \\
    & Numerical\_paraphrase & 224 & 0.58 (91)  & 0.74 (133) & 0.66 \\
    & Paraphrase & 1500 & 0.73 (750)  & 0.70 (750) & 0.72 \\
    & Text\_appended & 1500 & 0.57 (750)  & 0.70 (750) & 0.63 \\

    \bottomrule
  \end{tabular}%
}
  \centering
\caption{Intervention and Section-based results on test data using best model across both labels. Along with the $F_1$ score we also show number of instances.}
\label{tab:intervention_section_based_results}
\end{table*}

\subsection{Handling Long Premise-Hypothesis Pairs}
\label{app:handle_long_sequence}

\begin{table}[h]
\centering
\resizebox{\columnwidth}{!}{
\begin{tabular}{@{}lcccc@{}}
\toprule
\textbf{Model} & \textbf{Token Length} & \textbf{Mode} & \textbf{Dev $F_1$} \\
\midrule
Mistral-7B-v0.1 & 1024 & Remove & 0.71 \\
Mistral-7B-v0.1 & 1024 & Truncate & 0.72 \\
Mistral-7B-v0.1 & 2048 & Remove & 0.72 \\
Mistral-7B-v0.1 & 2048 & Truncate & \textbf{0.73} \\
\bottomrule
\end{tabular}%
}
\caption{Impact of different token length and strategy for handling long text}
\label{table:tok_len_stats}
\end{table}

One of the challenges of processing CTRs is their extensive length when paired up to form a premise-hypothesis pair. The Mistral model allows for token lengths of up to 4096. We experimented with different token lengths to see how they impacted the model's performance. We trained models with token lengths of 1024 and 2048 and evaluated their performance on the dev set. From \Cref{table:tok_len_stats}, we can see that increasing token length improved the results. We also tested the impact of truncating or removing text if it exceeded the token length. We observed that removing long text had a slight negative impact on the performance of the model. We used a token length of 4096 for our system development, with a truncation strategy in place for text that exceeds the token length limit.

\subsection{Dataset Difficulty Analysis Details}
\label{app:dataset_difficulty}

\begin{table*}[h]
  \small\centering
  \resizebox{\textwidth}{!}{
  \begin{tabular}{lcccccccc}

    \toprule

    {\bfseries Difficulty} & \multicolumn{2}{|c|}{{\bfseries Type}} & \multicolumn{4}{|c|}{{\bfseries Section}} & \multicolumn{2}{|c|}{{\bfseries Label}} \\
    \midrule\midrule[.1em]
     & & Count & \texttt{Intervention} & \texttt{Eligibility} & \texttt{Adverse Events} & \texttt{Results} & contradiction & entailment \\
    \midrule\midrule[.1em]

    \multirow{2}{*}{Easy}
    & \textit{single} & 234 & 50 & 132  & 8 & 44 & 61  & 173 \\
    & \textit{comparison} & 88 & 44 & 37  & 1 & 6 & 29  & 59 \\
    
    \multirow{2}{*}{Hard}
     & \textit{single} & 59 & 3 & 3  & 36 & 17 & 50  & 9 \\
    & \textit{comparison} & 37 & 11 & 4  & 20 & 2 & 29  & 8 \\

    \midrule[.1em] 

    \multirow{2}{*}{Easy-MinMax}
    & \textit{single} & 433 & 75 & 225  & 21 & 112 & 99  & 334 \\
    & \textit{comparison} & 237 & 112 & 93  & 5 & 27 & 66  & 171 \\
    
    \multirow{2}{*}{Hard-MinMax}
     & \textit{single} & 102 & 9 & 9  & 60 & 24 & 83  & 19 \\
    & \textit{comparison} & 88 & 17 & 10  & 54 & 7 & 66  & 22 \\

    \midrule[.1em] 

    \multirow{2}{*}{Easy-DataCartography}
    & \textit{single} & 465 & 88 & 155  & 128 & 94 & 233  & 232 \\
    & \textit{comparison} & 201 & 95 & 45  & 41 & 20 & 132  & 69 \\
    
    \multirow{2}{*}{Hard-DataCartography}
     & \textit{single} & 108 & 13 & 27  & 39 & 29 & 76  & 32 \\
    & \textit{comparison} & 71 & 24 & 24  & 21 & 2 & 46  & 25 \\

    \bottomrule
  \end{tabular}%
}
  \centering
\caption{Frequency of easy and hard examples across sections, instance type, and labels as identified by MinMax and data cartography methods. We also present combined results that is, the instances which are labeled easy and hard by both methods (Difficulty: Easy and Hard).}
\label{tab:data_difficulty_stats}
\end{table*}
For the MinMax weights approach, we first calculated the mean weight of correctly predicted instances. Every correctly predicted instance with a weight lower than the mean weight was selected as an easy instance (670). Similarly, for incorrectly predicted instances, we calculated their mean weight. Every incorrectly predicted instance with a weight higher than the mean weight was selected as a hard example (190). 

With the data cartography strategy, we calculated the mean confidence for correctly predicted instances. Every instance with a confidence higher than the mean confidence was considered an easy-to-learn example (666). Similarly, every incorrectly predicted instance with a confidence lower than the mean confidence of incorrectly predicted instances was considered hard to learn (179).

\begin{figure*}[h]
    \centering
    \includegraphics[width=\textwidth]{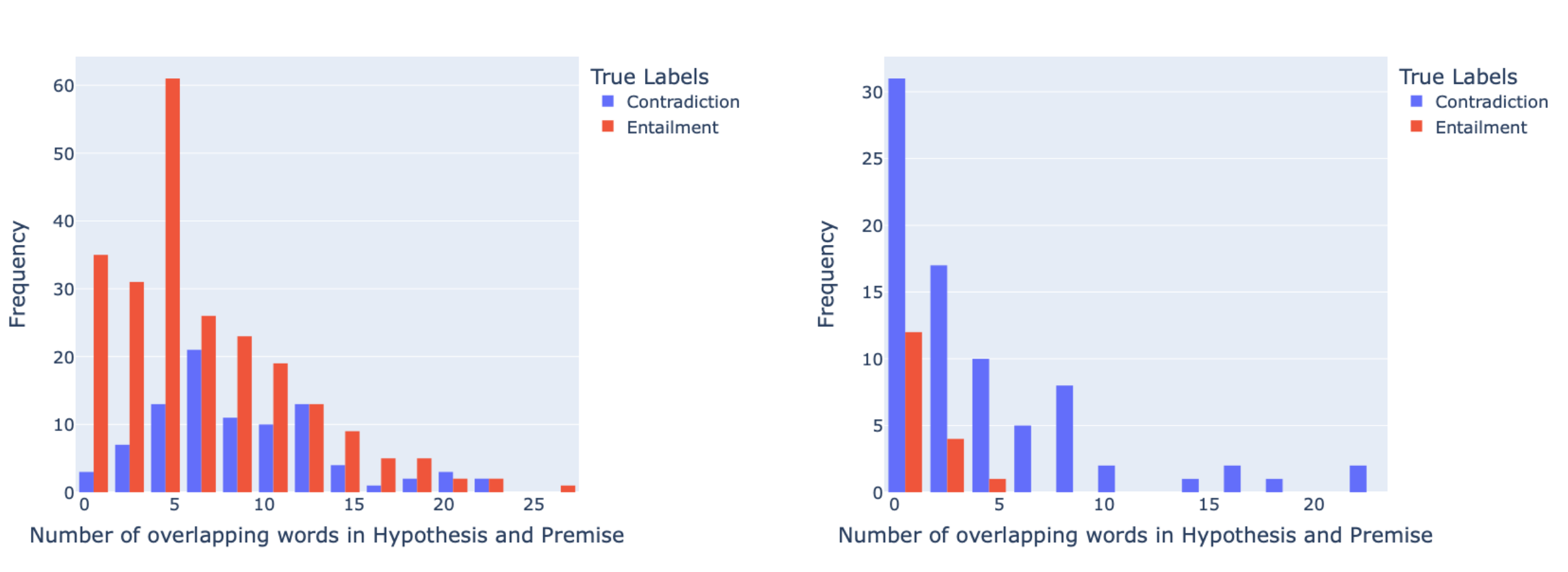}
    \centering
    \caption{Word overlap between hypothesis and premise with respect to true labels in Hard examples (\textbf{on the right}) and Easy examples (\textbf{on the left}).}
    \label{fig:label_overlap}
\end{figure*}

Furthermore, we manually examined four examples, two from each method Minmax and data-cartography labeled as most hard or difficult to learn. Three out of the four examples target the Adverse Events section, with one targeting the Results section. Notably, all four examples involved numerical reasoning, suggesting that the model still struggles with numerical reasoning despite demonstrating promising results on numerical interventions in the test data. For more details, see \Cref{tab:data_difficulty_stats}.

 A high overlap between the premise and hypothesis can lead to incorrect predictions of \textbf{entailment} relations, while low overlap can result in incorrect \textbf{contradiction} \citep{naik2018stress}. Analysis of the hard examples in \Cref{fig:label_overlap} revealed that instances with high overlap predominantly belong to \textbf{contradiction} relations, however, were incorrectly predicted as \textbf{entailment} relations by the model. This phenomenon could be attributed to the model associating higher word overlap with \textbf{entailment} relations, as evidenced by the easy examples in \Cref{fig:label_overlap}. However, such a correlation was not observed in the low word overlap region.

Using our trained model (\textit{MINMAX-MEDNLI-FT-NLI4CT}), we generated explanations alongside responses for each of these ten instances by increasing the number of generated tokens during the inference \footnote{This part was added after the first submission.}. As outlined in the work of \citet{swayamdipta2020dataset}, hard examples with low confidence scores may suggest mislabeled instances. We show two of these potential mislabeled instances in the \Cref{app:mislab}. Similarly, we also show the instances where the model confused cohorts and trials in \Cref{app:confused}.

\clearpage


\subsection{Potential Mislabeled Instances}
\label{app:mislab}
\includepdf[pages=-]{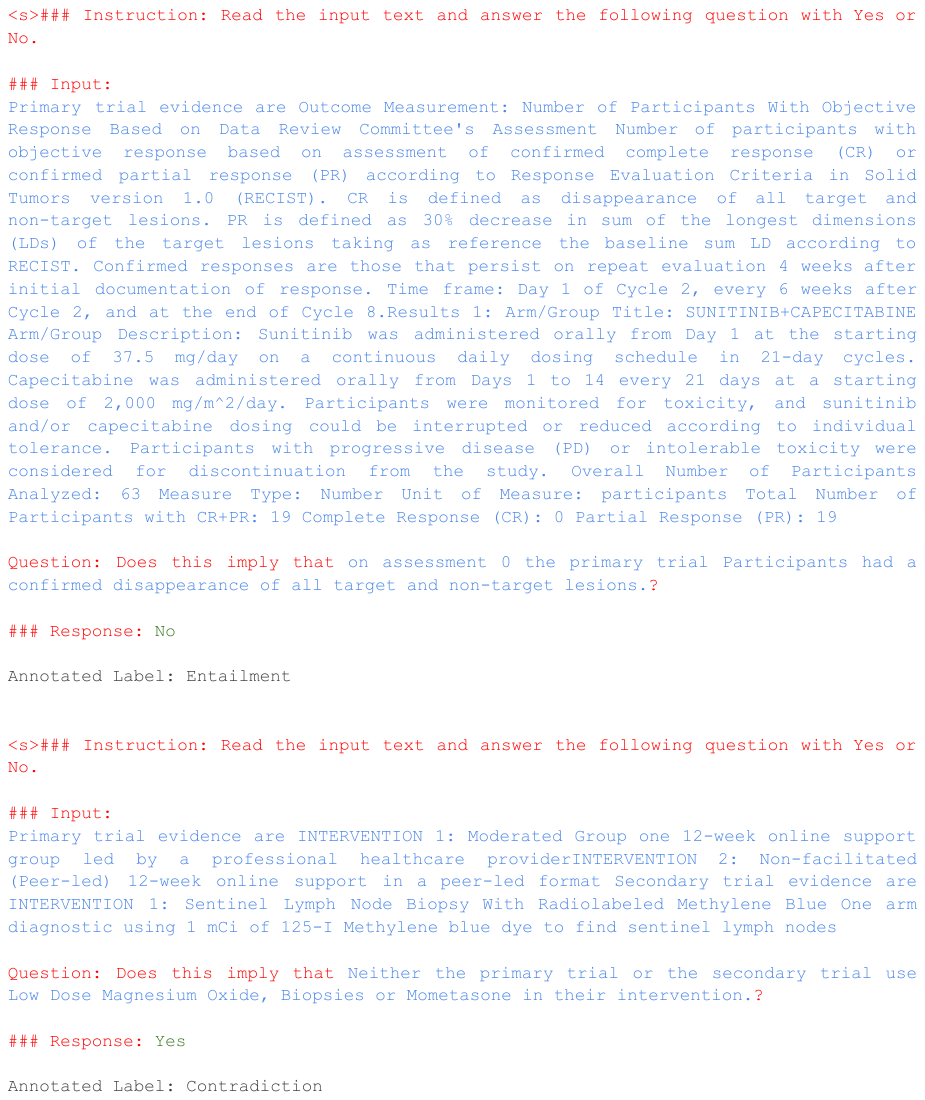}

\subsection{Cohorts and Trial Confusion}
\label{app:confused}


\section*{Supplementary material}

## A.9 Potential Mislabeled Instances

```
<s>### Instruction: Read the input text and answer the following question with Yes or No.

### Input:
Primary trial evidence are Outcome Measurement: Number of Participants With Objective Response Based on Data Review Committee's Assessment Number of participants with objective response based on assessment of confirmed complete response (CR) or confirmed partial response (PR) according to Response Evaluation Criteria in Solid Tumors version 1.0 (RECIST). CR is defined as disappearance of all target and non-target lesions. PR is defined as 30% decrease in sum of the longest dimensions (LDs) of the target lesions taking as reference the baseline sum LD according to RECIST. Confirmed responses are those that persist on repeat evaluation 4 weeks after initial documentation of response. Time frame: Day 1 of Cycle 2, every 6 weeks after Cycle 2, and at the end of Cycle 8.Results 1: Arm/Group Title: SUNITINIB+CAPECITABINE Arm/Group Description: Sunitinib was administered orally from Day 1 at the starting dose of 37.5 mg/day on a continuous daily dosing schedule in 21-day cycles. Capecitabine was administered orally from Days 1 to 14 every 21 days at a starting dose of 2,000 mg/m^2/day. Participants were monitored for toxicity, and sunitinib and/or capecitabine dosing could be interrupted or reduced according to individual tolerance. Participants with progressive disease (PD) or intolerable toxicity were considered for discontinuation from the study. Overall Number of Participants Analyzed: 63 Measure Type: Number Unit of Measure: participants Total Number of Participants with CR+PR: 19 Complete Response (CR): 0 Partial Response (PR): 19

Question: Does this imply that on assessment 0 the primary trial Participants had a confirmed disappearance of all target and non-target lesions.?

### Response: No

Annotated Label: Entailment


<s>### Instruction: Read the input text and answer the following question with Yes or No.

### Input:
Primary trial evidence are INTERVENTION 1: Moderated Group one 12-week online support group led by a professional healthcare providerINTERVENTION 2: Non-facilitated (Peer-led) 12-week online support in a peer-led format Secondary trial evidence are INTERVENTION 1: Sentinel Lymph Node Biopsy With Radiolabeled Methylene Blue One arm diagnostic using 1 mCi of 125-I Methylene blue dye to find sentinel lymph nodes

Question: Does this imply that Neither the primary trial or the secondary trial use Low Dose Magnesium Oxide, Biopsies or Mometasone in their intervention.?

### Response: Yes

Annotated Label: Contradiction
```

## A.10 Cohorts and Trial Confusion

### Instruction: Read the input text and answer the following question with Yes or No.

### Input:
Primary trial evidence are Adverse Events 1: Total: 4/42 (9.52%) Perforation, GI 1/42 (2.38%) Febrile neutropenia 1/42 (2.38%) Syncope 1/42 (2.38%) Rash/desquamation 1/42 (2.38%)

Question: Does this imply that 1/42 patients in cohort 2 of the primary trial fainted.?

### Response: Yes

Annotated Label: Contradiction

### Instruction: Read the input text and answer the following question with Yes or No.

### Input:
Primary trial evidence are Adverse Events 1: Total: 267/744 (35.89%) Neutropenia *2/744 (0.27%) Anaemia *1/744 (0.13%) Leukopenia *1/744 (0.13%) Thrombocytopenia *1/744 (0.13%) Thrombotic thrombocytopenic purpura *1/744 (0.13%) Atrial flutter *1/744 (0.13%) Cardiac arrest *1/744 (0.13%) Myocardial ischaemia *1/744 (0.13%) Arrhythmia *0/744 (0.00%) Cardiac failure congestive *0/744 (0.00%)Adverse Events 2: Total: 67/736 (9.10%) Neutropenia *1/736 (0.14%) Anaemia *0/736 (0.00%) Leukopenia *0/736 (0.00%) Thrombocytopenia *0/736 (0.00%) Thrombotic thrombocytopenic purpura *0/736 (0.00%) Atrial flutter *0/736 (0.00%) Cardiac arrest *0/736 (0.00%) Myocardial ischaemia *0/736 (0.00%) Arrhythmia *2/736 (0.27%) Cardiac failure congestive *1/736 (0.14%) Secondary trial evidence are Adverse Events 1: Total: 6 Atrial fibrillation 1/67 (1.49%) Ventricular fibrillation 1/67 (1.49%) Gastrointestinal perforation 1/67 (1.49%) Periproctitis 1/67 (1.49%) General physical health deterioration 1/67 (1.49%) Escherichia sepsis 1/67 (1.49%) Pneumonia 1/67 (1.49%) Tumour pain 1/67 (1.49%) Renal failure acute 1/67 (1.49%) Pleurisy 1/67 (1.49%)

Question: Does this imply that The most common adverse events in the primary trial and the secondary trial is Neutropenia with a total of 3 cases across all cohorts.?

### Response: No

Annotated Label: Entailment



\end{document}